%% file: paper.tex
\def\year{2022}\relax
\documentclass[letterpaper]{article} % DO NOT CHANGE THIS
\usepackage{aaai22}  % DO NOT CHANGE THIS
\usepackage{times}  % DO NOT CHANGE THIS
\usepackage{helvet}  % DO NOT CHANGE THIS
\usepackage{courier}  % DO NOT CHANGE THIS
\usepackage[hyphens]{url}  % DO NOT CHANGE THIS
\usepackage{graphicx} % DO NOT CHANGE THIS
\usepackage{natbib}  % DO NOT CHANGE THIS AND DO NOT ADD ANY OPTIONS TO IT
\usepackage{caption} % DO NOT CHANGE THIS AND DO NOT ADD ANY OPTIONS TO IT
\DeclareCaptionStyle{ruled}{labelfont=normalfont,labelsep=colon,strut=off} % DO NOT CHANGE THIS
\usepackage{algorithm}
\usepackage{algorithmic}

\usepackage{newfloat}
\usepackage{listings}
\floatstyle{ruled}
\newfloat{listing}{tb}{lst}{}
\floatname{listing}{Listing}

\usepackage{pgfplots}
\pgfplotsset{
tick label style = {font=\sansmath\sffamily},
every axis label/.append style={font=\sffamily\footnotesize},
compat=1.17,
}

\usepackage{placeins}

\nocopyright

\usepackage{acronym}

\usepackage[binary-units,separate-uncertainty]{siunitx}
\usepackage{xspace}
\usepackage{amsfonts}
\usepackage{amsmath}
\usepackage{amsthm}
\usepackage[mode=buildnew]{standalone}
\usepackage{tikz}
\usepackage[inline]{enumitem}
\usepackage{caption}
\usepackage{subcaption}
\usetikzlibrary{calc,through,backgrounds,shadows,decorations.pathreplacing}
\usetikzlibrary{positioning,decorations.text,arrows,patterns}
\usetikzlibrary{shapes,shapes.geometric,shapes.symbols}
\usetikzlibrary{arrows.meta}

\usepackage[textsize=scriptsize]{todonotes}
\usepackage{xcolor}
\input{clips.tex}

\newcommand*{\walle}{\textsc{Wall-E}\xspace}
\newcommand*{\rd}{\textsc{R2D2}\xspace}
\newcommand*{\eve}{\textsc{Eve}\xspace}
\newcommand*{\popf}{\textsc{POPF}\xspace}

\acrodef{RCLL}{RoboCup Logistics League}
\acrodef{CX}{CLIPS Executive}
\acrodef{MPS}{Modular Production System}
\acrodef{refbox}{referee box}
\acrodef{BS}{base station}
\acrodef{CS}{cap station}
\acrodef{RS}{ring station}
\acrodef{DS}{delivery station}
\acrodef{SS}{storage station}
\acrodef{TIL}{timed initial literal}
\acrodef{GDA}{goal-driven autonomy}
\acrodef{HGN}{Hierarchical Goal Network}
\acrodef{GLN}{Goal Lifecycle Network}
\acrodef{MRC}{Multi-Agent Coordination}
\acrodef{T-REX}{Teleo-Reactive Executive}
\newcommand*{\acfiu}[1]{\acfi{#1}\acused{#1}}
\newcommand*{\acfiup}[1]{\emph{\acl*{#1}}~(\acsp*{#1})\acused{#1}}

\newtheorem{definition}{Definition}

\newcommand*{\titletext}{Towards Using Promises for Multi-Agent Cooperation in Goal Reasoning}
\newcommand*{\authorstext}{Daniel Swoboda, Till Hofmann, Tarik Viehmann, Gerhard Lakemeyer}
\title{\titletext}
\author{\authorstext}
\affiliations{
    Knowledge-Based Systems Group\\
    RWTH Aachen University, Germany\\
    \{swoboda,hofmann,viehmann,lakemeyer\}@kbsg.rwth-aachen.de
}

\begin{document}

\maketitle
\begin{abstract}
  \input{abstract.txt}
\end{abstract}

\section{Introduction}

%Typically, research on intelligent agents is focused on goal-directed behavior,
%where an intelligent agent ``should be able to reason about actions in an
%autonomous manner in order to change the state of the world [...] as a means of
%satisfying a \emph{given} goal'' \cite{vattamBreadthApproachesGoal2013}. The
%goal itself is accepted as input by the user or inherent in the system.

While classical planning focuses on determining a plan that accomplishes a fixed goal, a goal
reasoning agent reasons about which goals to pursue and continuously refines its goals during
execution. It may decide to suspend a goal, re-prioritize its goals, or abandon a goal completely.
Goals are represented explicitly, including goal constraints, relations between multiple goals, and
tasks for achieving a goal.
Goal reasoning is particularly interesting on mobile robots,  as a mobile
robot necessarily acts in a dynamic environment and thus needs to be able to react to unforeseen
changes. Computing a single plan to accomplish an overall objective is often unrealistic, as
the plan becomes unrealizable during execution. While planning techniques such as contingent planning
\cite{hoffmannContingentPlanningHeuristic2005}, conformant planning
\cite{hoffmannConformantPlanningHeuristic2006}, hierarchical planning
\cite{kaelblingHierarchicalTaskMotion2011}, and continual planning
\cite{brennerContinualPlanningActing2009,hofmannContinualPlanningGolog2016} allow to deal with some
types of execution uncertainty, planning for the overall objective often simply is too time consuming. Goal
reasoning solves these problems by splitting the objective into smaller goals that can be
easily planned for, and allows to react to changes dynamically by refining the agent's goals.
This becomes even more relevant in multi-agent settings, as not only the environment is dynamic, but
the other agents may also act in a way that affects the agent's goals. While some work on
multi-agent goal reasoning exists
\cite{robertsCoordinatingRobotTeams2015,robertsGoalLifecycleNetworks2021,hofmannMultiagentGoalReasoning2021},
previous approaches focus on conflict avoidance, rather than facilitating active cooperation
between multiple agents.
%(e.g., two robots moving to the same location).
%rather than effective collaboration.
%Consider the two robots \walle and \rd from \todo{Add a figure
%showing the scenario}.  Their objective is to produce xenonite by processing raw materials in a
%multi-step process. \walle is currently at the first machine, which turns the raw material regolith
%into processite. \rd cannot use the same machine, as the machine's input is currently blocked by
%\walle. \rd also does not know the current goal of \walle, so it needs to wait until \walle has
%finished.

In this paper, we propose an extension to multi-agent goal reasoning that allows effective
collaboration between agents. To do this, we attach a set of \emph{promises} to a goal, which intuitively describe
a set of facts that will become true after the goal has been achieved. In order to make use of
promises, we extend the goal lifecycle \cite{robertsIterativeGoalRefinement2014} with \emph{goal
operators}, which, similar to action operators in planning,
provide a formal framework that defines when a goal can be formulated and which objective it pursues.
We use promises to evaluate the goal precondition not only
in the current state, but also in a time window of future states (with some fixed time horizon). This allows the
goal reasoning agent to formulate goals that are currently not yet achievable, but will be in the
future. To expand the goal into a plan, we then translate promises into \emph{timed initial
literals} from PDDL2.2 \cite{edelkampPDDL2LanguageClassical2004}, thereby allowing a PDDL planner to
make use of the promised facts to expand the goal.
With this mechanism, promises enable active collaborative behavior between multiple, distributed goal reasoning
agents.
%In the example above, \walle may promise that the machine will output processite in
%\SI{5}{\sec}. \rd can use this to formulate the goal to bring the processite from machine 1 to
%machine 2, even though the goal is not yet realizable. In effect, it can already move to the machine
%and wait for it to output the product, resulting in more effective collaboration.

The paper is organized as follows: In Section~\ref{sec:background}, we summarize goal reasoning and
discuss related work. As basis of our implementation, we summarize the main concepts of the \acf{CX}
in Section~\ref{sec:cx}.  In Section~\ref{sec:goal-preconditions}, we introduce a formal notion of a
goal operator, which we will use in Section~\ref{sec:promises} to define promises.  In
Section~\ref{sec:evaluation}, we evaluate our approach in a distributed multi-agent scenario, before
we conclude in Section~\ref{sec:conclusion}.

\section{Background and Related Work}\label{sec:background}
\paragraph{Goal Reasoning.}
In goal reasoning, agents can ``deliberate on and self-select their objectives''
\cite{ahaGoalReasoningFoundations2018}. Several kinds of goal reasoning have been proposed
\cite{vattamBreadthApproachesGoal2013}:
The \acfiu{GDA}
framework~\cite{munoz-avilaGoaldrivenAutonomyCasebased2010,coxModelPlanningAction2016} is based  on
finite state-transition systems known from planning \cite{ghallabAutomatedPlanningActing2016} and
define a goal as conjunction of first-order literals, similar to goals in classical planning. They
define a goal transformation function $\beta$, which, given the current state $s$ and goal $g$,
formulates a new goal $g'$. In the \ac{GDA} framework, the goal reasoner also produces a set of
\textit{expectations}, which are constraints that are predicted to be fulfilled during the
execution of a plan associated with a goal. In contrast to promises, which we also intend as
a model of constraints to be fulfilled, those
expectations are used for discrepancy detection rather than multi-agent coordination.
The \acfiu{T-REX} architecture \cite{mcgannDeliberativeArchitectureAUV2008} is a goal-oriented system that employs multiple
levels of reasoning abstracted in reactors, each of which operates in its own functional and
temporal scope (from the entire mission duration to second-level operations). Reactors on lower levels
manage the execution of subgoals generated at higher levels, working in synchronised timelines
which capture the evolution of state-variables over time. While through promises we also
attempt to encode a timeline of (partial) objective achievement, our approach has no hierarchy
of timelines and executives. Timing is not used to coordinated between levels of abstraction
on one agent, but instead is used to indicate future world-states to other, independent agents
reasoning in the same temporal scope.
%\showthe\textwidth
%
A \acfiu{HGN} \cite{shivashankarHierarchicalGoalbasedFormalism2012} is a partial order of goals,
where a goal is selected for execution if it has no predecessor. \acp{HGN} are used in the
\textsc{GoDeL} planning system \cite{shivashankarGoDeLPlanningSystem2013} to decompose a planning
problem, similar to HTN planning. \citet{robertsGoalLifecycleNetworks2021} extend \acp{HGN} to
\acfiup{GLN} to integrate them with the goal lifecycle.
The goal lifecycle~\cite{robertsIterativeGoalRefinement2014} models goal reasoning as an iterative
refinement process and describes how a goal progresses over time.
As shown in Figure~\ref{fig:cx-goal-lifecycle}, a goal is first \emph{formulated}, merely stating
that it may be relevant. Next, the agent \emph{selects} a goal that it deems the goal to be useful.
The goal is then \emph{expanded}, e.g., by querying a PDDL planner for a plan that accomplishes the
goal. A goal may also be expanded into multiple plans, the agent then \emph{commits} to one
particular plan.  Finally, it \emph{dispatches} a goal by starting to execute the plan.
It has been implemented in \emph{ActorSim} \cite{robertsActorSimToolkitStudying2016} and in the
\ac{CX} \cite{niemuellerGoalReasoningCLIPS2019}, which we extend in this work with promises.
%
%With few exceptions (e.g., \cite{hofmannMultiagentGoalReasoning2021}),
Most goal reasoning approaches focus on scenarios with single agents or scenarios where a central
coordinating instance is available. In contrast, our approach focuses on a distinctly decentralized
multi-agent scenario with no central coordinator.

\paragraph{Multi-Agent Coordination.}
In scenarios where multiple agents interact in the same environment, two or more agents might
require access to the same limited resources at the same time. In such cases, multi-agent
coordination is necessary to avoid potential conflicts, create robust solutions and enable
cooperative behavior between the agents \cite{diasMarketBasedMultirobotCoordination2006}.
%
%Because of the diverse range of agent setups and multitude of settings in which multi-agent
%systems are used, various coordination strategies have been proposed, a sample is given here.
Agents may coordinate implicitly by recognizing or learning a model of another agent's behavior
\cite{sukthankarPlanActivityIntent2014,albrechtAutonomousAgentsModelling2018}.
Particularly relevant is \emph{plan recognition}~\cite{carberryTechniquesPlanRecognition2001}, where
the agent tries to recognize another agent's goals and actions.
\textsc{Alliance}~\cite{parkerALLIANCEArchitectureFault1998} also uses an implicit coordination
method based on \emph{impatience} and \emph{acquiescence}, where an agent eventually takes over a
goal if no other agent has accomplished the goal (impatience), and eventually abandons a goal if it
realizes that it is making little progress (acquiescence).
In contrast to such approaches, we propose that the agents explicitly communicate (parts of) their
goals, so other agents may directly reason about them.
To obtain a conflict-free plan, each agent may start with its own individual plan and iteratively
resolve any flaws~\cite{coxEfficientAlgorithmMultiagent2005}, which may also be modeled as a
distributed constraint optimization problem~\cite{coxDistributedFrameworkSolving2005}.
Instead of starting with individual plans, \citet{jonssonScalingMultiagentPlanning2011} propose to
start with some initial (sub-optimal) shared plan and then let each agent improve its own actions.
Alternatively, agents may also \emph{negotiate} their goals
\cite{davisNegotiationMetaphorDistributed1983,krausAutomatedNegotiationDecision2001}, e.g., using
\emph{argumentation}~\cite{krausReachingAgreementsArgumentation1998}, which typically requires an
explicit model of the mental state.
\citet{vermaMultiRobotCoordinationAnalysis2021}
classify coordination mechanisms based on properties such as static vs dynamic, weak vs strong,
implicit vs explicit, and centralized vs decentralized.
Here, we focus on dynamic, decentralized and distributed multi-robot coordination.
Alternatively, coordination approaches can be classified based on organization structures
\cite{horlingSurveyMultiagentOrganizational2004}, e.g., coalitions, where agents cooperate but each
agent attempts to maximize its own utility, or teams, where
the agents work towards a common goal.
Such teams may also form dynamically, e.g., by dialogues~\cite{dignumAgentTheoryTeam2001}.
Here, we are mainly interested in fixed teams, where a fixed group of robots have a common goal.
Role assignment approaches attempt to assign fixed and distinct roles to each of the
participating agents and thereby fixing the agent's plan.
This can be achieved either by relying on a central instance or through distributed approaches~
\cite{iocchiDistributedCoordinationHeterogeneous2003,vailDynamicMultiRobotCoordination2003,jinDynamicTaskAllocation2019}.
Intention sharing approaches allow agents to incorporate the objectives of other agents
into their own reasoning in an effort to preemptively avoid conflicting actions~\cite{holvoetBeliefsDesiresIntentions2006,grantLogicBasedModelIntentions2002,sarrattPolicyCommunicationCoordination2016}.
\emph{SharedPlans}~\cite{groszCollaborativePlansComplex1996} use complex actions that involve
multiple agents.  They use an intention sharing mechanism, formalize collaborative
plans, and explicitly model the agents' mental states, forming mutual beliefs.
Similar to promises, an agent may share its intention with \emph{intending-that} messages, which
allow an agent to reason about another agent's actions.
Depending on the context, agents must have some form of
\emph{trust}~\cite{yuSurveyMultiAgentTrust2013,pinyolComputationalTrustReputation2013} before they
cooperate with other agents.
Market-based approaches as surveyed by \citet{diasMarketBasedMultirobotCoordination2006}
use a bidding process to allocate different conflict-free tasks between the agents of a multi-robot system.
This can be extended to auction of planning problems in an effort to optimize temporal multi-agent
planning~\cite{hertleEfficientAuctionBased2018}.
In the context of multi-agent goal reasoning, \citet{wilsonGoalReasoningModel2021} perform
the goal lifecycle for an entire multi-robot system on each agent separately and then
use optimization methods to prune the results.
%The \ac{CX} also provides 
%\todo{maybe move to CX section?} employs two mechanisms for coordination in a multi-agent setup: locking actions,
%which operate on an action level, and goal resources which operate on a goal level.
%Locking actions prevent conflict situations in critical parts of an action sequence. Goal
%resources can be used to prevent two agents from dispatching goals that want to operate on
%the same limited resources.

% Classification of our own approach according to \cite{vermaMultiRobotCoordinationAnalysis2021}
%dynamic, weak (where promises are a simple form of strong coordination), mostly implicit,
%decentralized, distributed;
%useful for multi-robot systems that are coordinated, heterogeneous or homogeneous, cooperative

\section{The CLIPS Executive}\label{sec:cx}

The \acf{CX}~\cite{niemuellerGoalReasoningCLIPS2019} is a goal reasoning system implemented in the
CLIPS rule-based production system \cite{wygantCLIPSPowerfulDevelopment1989}. Goals follow the
goal lifecycle \cite{robertsIterativeGoalRefinement2014}, with domain specific rules guiding
their progress. It uses PDDL to define
a domain model, which is parsed into the CLIPS environment to enable continuous reasoning on the
state of the world, using a representation that is compatible with a planner. Therefore, the notions
of plans and actions are the ones known from planning with PDDL. Multiple instances
of the \ac{CX} (e.g., one per robot), can be coordinated through locking and
domain sharing mechanisms, as demonstrated in the \acf{RCLL} \cite{hofmannMultiagentGoalReasoning2021}.
However, these mechansims only serve to avoid conflicts as each agent still reasons independently,
without considering the other agents intents and goals.

\paragraph{Goals.}
Goals are the basic data structure which model certain objectives to be achieved. 
Each goal comes with a set of preconditions which 
must be satisfied by the current world state before being formulated. In the \ac{CX}, these
preconditions are modelled through domain-specific CLIPS rules.

Each instance of a goal belongs to a class, which represents a certain objective
to be achieved by the agent. Additionally, the goal can hold parameters that might be required
for planning and execution. A concrete \emph{goal} is obtained by grounding all parameters of a goal class.
Depending on the current state of the world, each robot may formulate multiple goals and multiple
instances of the same goal class.

Each goal follows a \emph{goal lifecycle}, as shown in Figure~\ref{fig:cx-goal-lifecycle}.
The agent formulates a set of goals based on the current state of the world. These can
be of different classes, or multiple differently grounded versions of the same class. After
formulation, one specific goal is selected, e.g., by prioritizing goals
that accomplish important objectives. The selection mechanism may be modelled in
a domain-specific way, which allows adapting the behavior according to the domain-specific requirements.
In the next step, the selected goal is \emph{expanded} by
computing a plan with an off-the-shelf PDDL planner. Alternatively,
a precomputed plan can be fetched from a plan library. Once a plan
for a goal is determined, the executive attempts to acquire its required
resources. Finally, the plan is executed on a per-action basis.
%Plans can be sequential or temporal.
%In this work, we focus on sequential plans.

%Generally, goals are expected to be modelled in a per-domain basis. I.e., goals should be
%designed s.t. they describe objectives that the agent should achieve in certain states in
%the given domain. Compared to actions in planning, a goal in goal reasoning does not describe
%what needs to be changed in the world, but how the world needs to be changed.

\paragraph{Goals Example. }
To illustrate goals and goal reasoning in the CX, let us consider a simple resource mining scenario which we will
evolve throughout this paper and finally use as the basis for our conceptual evaluation. Different case examples 
based on the given setting are presented in Figure~\ref{fig:timeline-promises}.
Up to two robots (\walle and \rd) produce the material \emph{Xenonite} by first mining \emph{Regolith} in a mine, 
refining the Regolith to \emph{Processite} in a refinery machine, and
then using Processite to produce Xenonite in a production machine. Let us assume the following setting as the
basis for our example:
one machine is filled with Regolith, one machine is empty, the robots currently pursue no goals. 
In this scenario, each robot may pursue five classes of goals:
\begin{enumerate*}[label=(\alph*)]
  \item \textsc{FillContainer} to fill a container with Regolith at the mine or with Processite,
  \item \textsc{CleanMachine} to fill a container with Processite or Xenonite at a machine,
  \item \textsc{Deliver} a container to a machine,
  \item \textsc{StartMachine} to use a refinery or production machine to process the material,
  \item \textsc{DeliverXenonite} to deliver the finished Xenonite to the storage.
\end{enumerate*}
Because one machine is filled, a goal of class \textsc{StartMachine} can be formulated. The parameters of the
goal class are grounded accordingly, i.e., the target machine is set to the filled machine. Should both machines 
be filled at the same time, two goals of class \textsc{StartMachine} could be formulated, one for each of the machines.
The agent now selects one of the formulated goals for expansion. Since in our case only one goal is formulated,
it will be selected. Should there be \textsc{StartMachine} goals for each machine, then, e.g., the one operating on 
second machine in the production chain might be prioritized. 

\paragraph{Execution Model.}
After a goal has been expanded (i.e., a plan has been generated),
the \ac{CX} commits to a certain plan and then dispatches the goal
by executing the plan.
For sequential plans, whenever no action of the plan is executed, the \ac{CX} selects the next
action of the plan.
It then checks whether the action is executable by evaluating the action's precondition, and if so,
sends the action to a lower level execution engine.
If the action's precondition is not satisfied, it remains in the pending state until either the
precondition is satisfied or a timeout occurs. In a multi-agent scenario, each \ac{CX} instance
performs its own planning and plan execution independently of the other agents.

The execution of two goals with one robot in our example scenario is visualized in with one robot is visualized in Figure \ref{fig:timeline-promises} 
as \textsc{Scenario 1}. After finishing the goal of class \textsc{StartMachine}, a goal of class \textsc{CleanMachine} 
is formulated and dispatched, clearing the machine's output. 

\paragraph{Multi-Agent Coordination.}
The \ac{CX} supports two methods for multi-agent coordination: \emph{locking actions} acquire a lock
for exclusive access, e.g., a location so no two robots try to drive to the same location.
They are part of a plan and executed similar to regular actions.
\emph{Goal resources} allow coordination on the goal rather than the action level.
Each goal may have a set of \emph{required resources} that it needs to hold exclusively while the
goal is dispatched. If a second goal requires a resource that is already acquired by another goal,
the goal is rejected. This allows for coordination of conflicting goals, e.g., two robots picking up the
same container. Both methods are intended to avoid conflicts and overlapping goals between the agents,
rather than fostering active cooperation.

To illustrate coordination by means of goal resources, let us extend the previous example \textsc{Scenario 1} 
by adding the second robot. In \textsc{Scenario 2}, illustrated in Figure \ref{fig:timeline-promises}, both 
\walle and \rd formulate and select \textsc{StartMachine} for \texttt{M1}, 
which requires the machine as a resource.
\walle starts formulating first and manages to acquire the resources. Therefore it is able to dispatch its goal.
\rd on the other hand must reject the goal as it is not able to acquire the machine resource.
After its goal has been rejected, it may select a different goal. Since now the machine is occupied,
it has to wait until the preconditions of some class are met and a new goal can be formulated.
This is the case once the effects of \textsc{StartMachine} are
applied, which causes \rd to formulate and dispatch \textsc{CleanMachine} to clear the machine.

\paragraph{Execution Monitoring.}
The \ac{CX} includes execution monitoring functionality to handle exogenous events and action failure
\cite{niemuellerCLIPSbasedExecutionPDDL2018}. Since most actions interact with other agents or
the physical world, it is critical to monitor their execution for failure or delays. 
Action may either be retried or failed, depending on domain-specific evaluations of the execution behavior. 
Additionally, timeouts are used to detect potentially stuck actions. The failure of an action might lead to reformulation
of a goal. This means that the plan of the goal might be adapted w.r.t. the partial
change in the world state since original formulation. Alternatively, a completely new objective might be followed by the agent.

\section{Goal Reasoning with Goal Operators}\label{sec:goal-preconditions}
\begin{figure}[htb]
  \centering
  \includestandalone[width=1.0\linewidth]{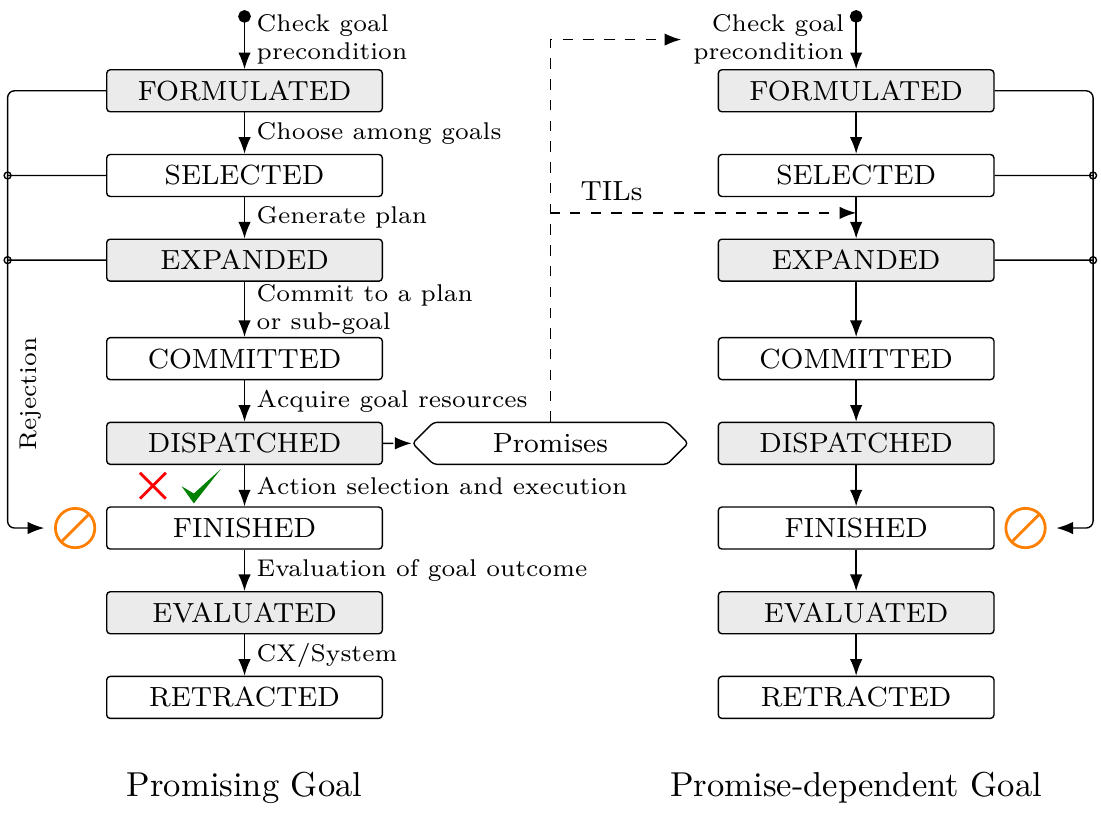}
  \caption{
    The \ac{CX} goal lifecycle \cite{niemuellerGoalReasoningCLIPS2019} of two goals and their
    interaction by means of promises.
    When the first goal is dispatched, it makes a set of promises, which can be used by the second
    goal for its precondition check.
    If the goal precondition is satisfied based on the promises, the goal can already be formulated
    (and subsequently selected, expanded, etc.) even if the precondition is not satisfied yet.
    Additionally, promises are also used as timed initial literals (TILs) for planning in order to
    expand a goal.
    %Once a goal is \emph{dispatched}, a set of literals are promised for
    %some future timepoint. These promises may be used to formulate another goal (possibly by a
    %different agent) that would otherwise not be possible yet.
}
  \label{fig:cx-goal-lifecycle}
\end{figure}

To define promises and how promises affect goal formulation in the context of goal reasoning,
we first need to formalize goals.
%Intuitively, a goal describes the intention of an agent to pursue a certain objective.
Similar to \citet{coxModelPlanningAction2016}, we base our definition of a goal on classical
planning problems.
However, as we need to refer to time later, we assume that each state is timed.
Formally, given a finite set of logical atoms $A$, a state is a pair $\left(s, t\right) \in 2^A
\times \mathbb{Q}$, where $s$ is the set of logical atoms that are currently true, and $t$ is the
current global time.
Using the closed world assumption, logical atoms not mentioned in $s$ are implicitly considered to
be false.
A literal $l$ is either a positive atom (denoted with $a$) or a negative atom (denoted with
$\overline{a}$).
For a negative literal $l$, we also denote the corresponding positive literal with $\overline{l}$
(i.e., $\overline{\overline{l}} = l$).
A set of atoms $s$ satisfies a literal $l$, denoted with $s \models l$ if
\begin{enumerate*}[label=(\arabic*)]
  \item $l$ is a positive literal $a$ and $a \in s$, or
  \item $l$ is a negative literal $\overline{a}$ and $a \not\in s$.
\end{enumerate*}
Given a set of literals $L = \{ l_1, \ldots, l_n \}$, the state $s$ satisfies $L$, denoted with $s
\models L$, if $s \models l_i$ for each $l_i \in L$.
%Using the usual connectors $\neg, \wedge, \vee$ known from propositional logic, we can construct propositional formulas over $A$.

We define a \emph{goal operator} similar to a planning operator in classical planning~
\cite{ghallabAutomatedPlanningActing2016}:

\begin{definition}[Goal Operator]
  A \emph{goal operator} is a tuple
  $\gamma = \left(\mathrm{head}(\gamma), \mathrm{pre}(\gamma), \mathrm{post}(\gamma)\right)$, where
  \begin{itemize}
    \item $\mathrm{head}(\gamma)$ is an expression of the form $\mathit{goal}(z_1, \ldots, z_k)$, where
      $\mathit{goal}$ is the \emph{goal name} and $z_1, \ldots, z_k$ are the \emph{goal parameters}, which
      include all of the variables that appear in $\mathrm{pre}(\gamma)$ and
      $\mathrm{post}(\gamma)$,
    \item $\mathrm{pre}(\gamma)$ is a set of literals describing the condition when a
      goal may be formulated,
    \item $\mathrm{post}(\gamma)$ is a set of literals describing the objective that the
      goal pursues, akin to a PDDL goal.
  \end{itemize}
\end{definition}

%Goal operators resemble concepts from different planning approaches. Landmarks \cite{hoffmannOrderedLandmarksPlanning2004}
%are facts that must be true at some point during the execution of any solution plan for a goal.
%Instead, goal postconditions ($\mathrm{post}(\gamma)$) are facts that must be true at the \textit{end}
%of any solution plan for a goal.
%Unlike effects of macro-actions \cite{colesMarvinHeuristicSearch2007},
%which are chains of compatible plan actions combined into a single action, $\mathrm{post}(\gamma)$
%is a partial description of the world state that we want to achieve. A goal does not dictate through
%which means (i.e. which actions) this state must be achieved. Instead, just like a goal state definition
%in planning, $\mathrm{post}(\gamma)$ just describes which facts must be true
%for an instance of $\gamma$ to be achieved. $\mathrm{pre}(\gamma)$ is not a precondition for
%a plan, action, or macro action which achieves $\gamma$, but a condition for the goal reasoner
%to indicate states, in which we think it is reasonable to consider an instance of this goal
%operator.

A goal is a ground instance of a goal operator.
A goal $g$ can be \emph{formulated} if $s \models \mathrm{pre}(g)$. When a goal is
finished, its post condition holds, i.e., $s \models \mathrm{post}(g)$.
Note that in contrast to an action operator (or a macro operator), the effects of a goal are not completely determined; any
sequence of actions that satisfies the objective $\mathrm{post}(g)$ is deemed feasible and
additional effects are possible.
Thus, the action sequence that accomplishes a goal is not pre-determined but computed by a planner
and may depend on the current state $s$.
Consider the goal \textsc{CleanMachine} from Listing \ref{lst:goal-operator}. Its objective defines
that the robot carries a container and that the container is filled.  However, to achieve this
objective given the set preconditions, the robot also moves away from its current location,
which will be an additional effect of any plan that achieves the goal.

%The goal reasoner then \emph{formulates} a goal $\mathit{goal}(z_1, \ldots, z_k\right)$ whenever the
%goal's precondition $\mathrm{pre}(\gamma)$ is satisfied in the current state, and then follows the
%goal life cycle as shown in Figure~\ref{fig:cx-goal-lifecycle}.

\begin{lstlisting}[style=ReallySmallCLIPSFloat,label={lst:goal-operator},
caption={%
The definition of the goal operator \textsc{Clean-Machine}. The precondition defined when a goal may be
formulated; the objective defines the PDDL goal to plan for once the goal has been selected.}]
(goal-operator (class CleanMachine)
  (param-names     r     side machine  c    mat)
  (param-types     robot loc  machine  cont material)
  (param-quantified)
  (lookahead-time 20)
  (preconditions "
      (and
          (robot-carries ?r ?c)
          (container-can-be-filled ?c)
          (location-is-free ?side)
          (location-is-machine-output ?side)
          (location-part-of-machine ?side ?machine)
          (machine-in-state ?machine READY)
          (machine-makes-material ?machine ?mat)
          (not (storage-is-full))
      )
  ")
  (objective 
  "(and (robot-carries ?r ?c)
        (container-filled ?c ?mat)
  )")
)
\end{lstlisting}
%\todo[inline]{Adapt implementation to actually use the slot 'objective'}

We have extended the \ac{CX} with goal operators (shown in
Listing~\ref{lst:goal-operator}). For each goal operator and every possible grounding of the
operator, the goal reasoner instantiates the corresponding goal precondition and tracks its
satisfaction while the system is evolving. Once the goal precondition is satisfied, the goal is
formulated. Afterwards, the goal follows the goal lifecycle as before
(Figure~\ref{fig:cx-goal-lifecycle}).

%\begin{itemize}
%  \item so far, goal formulation was unspecified and domain-dependent
%  \item now: define a succinct condition when some goal can be formulated
%  \item base the condition on PDDL formulas
%  \item the goal precondition can use the parameters of the goal and can use quantifiers
%  \item the formula is continuously evaluated by the executive
%  \item quantifiers are grounded by instantiating the sub-formula for each object instance
%  \item if a new object appears, the formula is extended accordingly
%  \item once the condition is satisfied, the corresponding goal is automatically formulated
%  \item formulation criterion is taken from \emph{goal operators}, somewhat similar to macro
%    operators
%  \item this then follows the goal life cycle as before, i.e., if the goal is selected (domain-dependent criterion), the goal is expanded, e.g., by calling a PDDL planner or by fetching a plan from the plan database
%\end{itemize}

\section{Promises}\label{sec:promises}

\begin{figure*}[p]
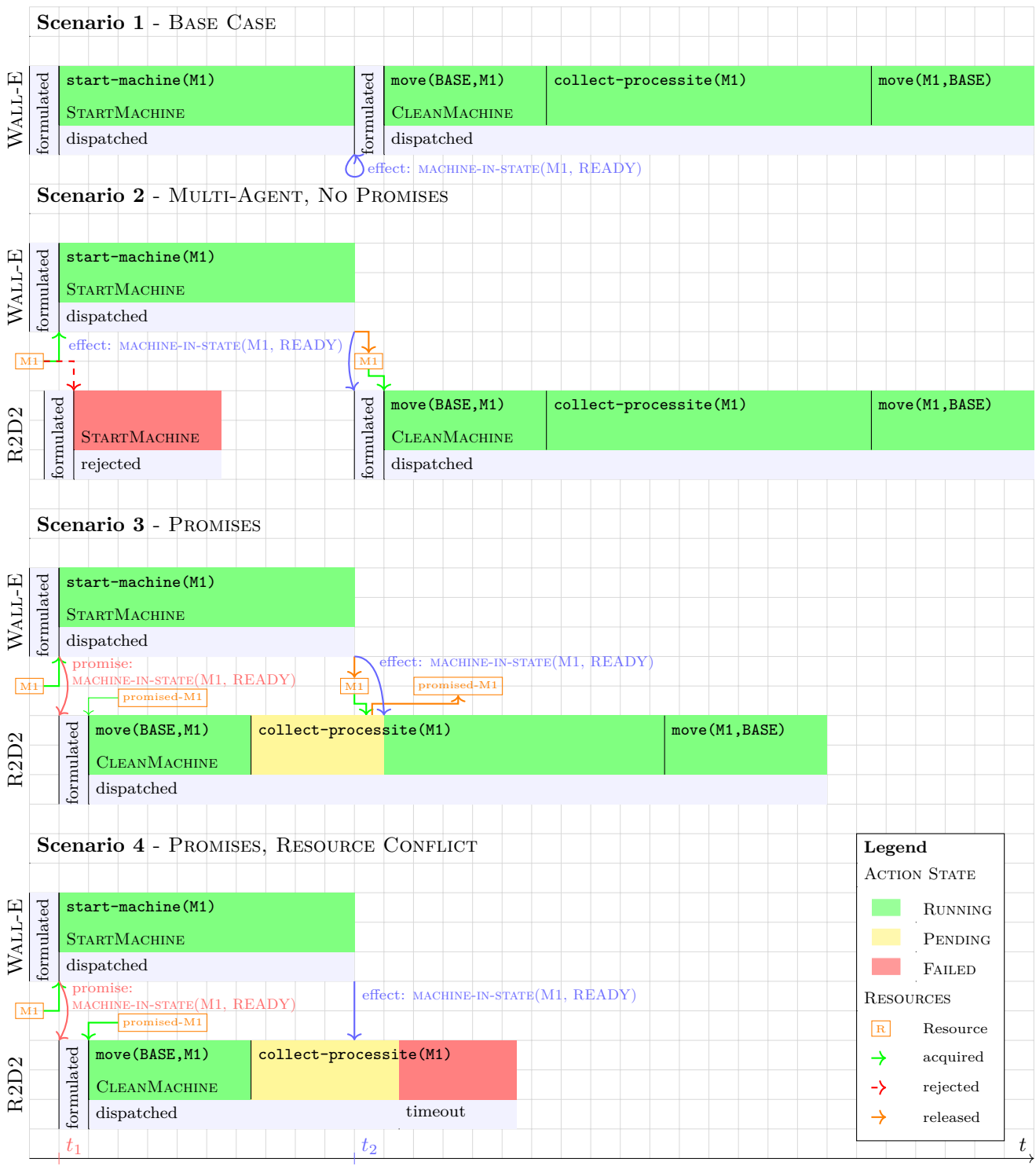

  %for some reason the figure is shifted right when included as standalone
  \centering
  \includestandalone{figures/scenarios}

  \caption{
    Multiple scenarios of interaction between two goals and robots in the \ac{CX} with and without promises
    as discussed in the running example.
    \textsc{Scenario 1}: one robot \walle and no promises. It successively formulates and executes
    the applicable goals.
    \textsc{Scenario 2}: two robots (\walle and \rd) and no promises. Both attempt to 
    dispatch \textsc{StartMachine}, illustrating interaction between multiple agents and goal resources. 
    \textsc{Scenario 3}: two robots, promises. \rd formulates \textsc{CleanMachine}
    earlier from the promises, but the action \texttt{collect-processite} is pending until
    the effects from \walle are applied. 
    \textsc{Scenario 4}: two robots, promises, and failed resource handover. \texttt{collect-processite} is pending
    until the resource \texttt{M1} is handed over. A timeout occurs and the goal fails.
  }
  \label{fig:timeline-promises}
\end{figure*}

\newcommand*{\promfrom}{\ensuremath{\mathtt{From}}}
\newcommand*{\promuntil}{\ensuremath{\mathtt{Until}}}

A goal reasoning agent relying on goal operators has a model of the conditions that have to be met
before it executes a goal $g$ ($\mathrm{pre}(g)$) and a partial model of the world
after the successful execution of its plan (its objective $\mathrm{post}(g)$). Thus, when the agent
dispatches $g$, it assumes that it will accomplish the objective at some point
in the future (i.e. $s \models \mathrm{post}(g)$). However, the other agents are oblivious
to $g$'s objectives and the corresponding changes to the world and therefore need to wait until the
first agent has finished its goal. To avoid this unnecessary delay, we introduce \emph{promises},
which intuitively give some guarantee for certain literals to be true in a future state:
\begin{definition}[Promise]
A \emph{promise} is a pair $\left(l, t\right)$, stating that the literal $l$ will be satisfied at
(global) time $t$.
\end{definition}
%(i.e., the atom will be true if $l$ is a positive literal and
%false if $l$ is a negative literal)

Promises are made to the other agents by a \emph{promising} agent when it dispatches a goal.
The \emph{receiving} agent may now use those promises to evaluate whether a goal can be formulated, even
if its precondition is not satisfied by the world.
More precisely, given a set of promises, we can determine the point in time when a goal precondition
will be satisfied:
%Promises determine the time when a formula is satisfied:
\begin{definition}[Promised Time]
Given a state $\left(s, t\right)$ and a set of promises $P = \{ \left(l_i, t_i\right) \}_i$, we define
$\promfrom(l, s, t, P)$ as the timepoint when a literal $l$ is satisfied and $\promuntil(l, s, t,
P)$ as the timepoint when $l$ is no longer satisfied:
\begin{align*}
  \promfrom (l, s, t, P) &=
  \begin{cases}
    t & \text{ if } s \models l
    \\
    \min_{\left(l, t_i\right) \in P} t_i & \text{ if } \exists t_i: \left(l, t_i\right) \in P
    \\
    \infty & \text{ else }
  \end{cases}
  \\
  \promuntil (l, s, t, P) &=
  \begin{cases}
    t & \text{ if } s \models \overline{l}
    \\
    \min_{\left(\overline{l}, t_i\right) \in P} t_i & \text{ if } \exists t_i: \left(\overline{l}, t_i\right) \in P
    \\
    \infty & \text{ else }
  \end{cases}
\end{align*}

We extend the definition to a set of literals $L$:
\begin{align*}
  \promfrom(L, s, t, P) &= \max_{ l_i \in L} \promfrom(l_i, s, t, P)
  \\
  \promuntil(L, s, t, P) &= \min_{l_i \in L} \promuntil(l_i, s, t, P)
\end{align*}
\end{definition}

Promises are tied to goals and can either be based on the effects of the concrete plan actions
of an expanded goal, or simply be based on a goal's postcondition. They can be extracted automatically
from these definitions, or hand-crafted in simple domains. Though the concept can be applied more
flexibly (e.g. dynamic issuing of promises based on active plans and world states), in our
implementation promises are static and created during goal expansion.

\subsection{Goal Formulation with Promises}
We continue by describing how promises and promise times for literals can be used for goal
formulation.
%Next, we consider promises during goal formulation.
As shown in Listing~\ref{lst:goal-operator}, we
augment the goal operator with a \emph{lookahead time}, which is used to evaluate the goal
precondition for future points in time. Given a set of promises $P$, a lookahead time $t_l$ and a goal precondition
$\mathrm{pre}(\gamma)$, the goal reasoner formulates a goal in state $\left(s, t\right)$ iff the
precondition
will be satisfied within the next $t_l$ time units, i.e., iff
%and additionally, the
%precondition is promised to remain satisfied for at least $t_l$ timesteps, i.e. iff:
\[
  %\left(s, t\right) \models
  \promfrom(\mathrm{pre}(\gamma),s, t, P) \leq t + t_l
\]
Note that if $s \models \mathrm{pre}(\gamma)$, then $\promfrom(\mathrm{pre}(\gamma), s, t, P) = t$ and thus
the condition is trivially satisfied.
Also, if the lookahead time is set to $t_l = 0$, then the goal is formulated iff its precondition is
satisfied in the current state, thus effectively disabling the promise mechanism.

Furthermore, this condition results in an optimistic goal formulation, as the
goal will still be formulated even if the goal's precondition is promised to be no longer satisfied
at a future point in time.
To ensure that a goal is only formulated if its precondition is satisfied for the whole lookahead
time, we can additionally require:
\[
  \promuntil(\mathrm{pre}(\gamma), s, t, P) \geq t + t_l
\]
%with $\promuntil(\mathrm{pre}(\gamma), s, t, P) = \infty$ if $s \models \mathrm{pre}(\gamma)$ and
%for any $l \in \mathrm{pre}(\gamma)$ and $t'$, $(\overline{l}, t') \not\in P$.
In our implementation, $\promuntil$
is currently not considered, as we are mainly interested in optimistic goal formulation.

Incorporating promises into goal formulation leads to more cooperative behavior
between the agents as goals that build on other goals can be considered by an agent
during selection, thereby enabling parallelism.
Thus, it is a natural extension of the goal lifecycle with an intention sharing mechanism.
%It incorporates the forsight of intention sharing approaches, as natural extension of the
%goal lifecycle. \todo{This is a very strong statement but
%I'm full of reviewers, remove or weaken as needed}

\paragraph{Promises in the CX.}
In the \ac{CX}, promises are continously evaluated to compute $\promuntil$ and $\promfrom$ for each
available grounding of a goal precondition formula. The results are stored for each
formula and computed in parallel to the normal formula satisfaction evaluation.
Therefore, promises do not directly interfere with the world model, but rather are
integrated into the goal formulation process by extending the precondition check to
also consider $\promfrom$ for a certain lookahead time $t_l$.

The lookahead time is manually chosen based on the expected time to accomplish the objective and are
specific for each goal. In scenarios with constant execution times for actions, those can be
used directly as the estimate. If execution times for actions are not constant
(e.g. driving actions with variable
start and end positions), average recorded execution times or more complex estimators might be used,
at the loss of promise accuracy. By choosing a lookahead time of $0$, promises can be ignored.

Promises are shared between the agents through the shared world model of the \ac{CX}. For now,
we hand-craft promises for each goal operator $\gamma$ based on $\mathrm{post}(g)$. However,
extracting promises automatically from postconditions and plans may be considered in future work.
If a goal is completed or fails, the promises are removed from the world
model.
%\begin{itemize}
%  \item Each goal can \emph{promise} literals for certain time points in the future,
%    \emph{promising} that the goal will accomplish those effects
%  \item Other agents can formulate goals based on these promises
%  \item In particular, in addition to evaluating whether a goal precondition is currently satisfied,
%    we also evaluate whether a goal precondition will be satisfied in the future
%  \item This is also tracked continuously in the executive
%  \item atomic sub-formulas take the time from the respective promise
%  \item conjunctions take the max, disjunctions the min
%  \item we can also skip formulating goals if the negated precondition is $\promfrom$
%  \item the definition makes the simplifying assumption that we cannot multiple positive and
%    negative promises for the same atom
%  \item planning with promises is planning with TILs
%\end{itemize}

%\begin{lstlisting}[style=ReallySmallCLIPSFloat]
%(deftemplate domain-promise
%  "A promise is a literal that will become true
%   in the future."
%  (slot name (type SYMBOL))
%  (multislot param-values)
%  (slot negated (type SYMBOL)
%                (allowed-values TRUE FALSE))
%  (slot promising-goal (type SYMBOL))
%  (slot promising-agent (type SYMBOL) (default nil))
%  (slot active (type SYMBOL) (default FALSE)
%               (allowed-values TRUE FALSE))
%  ; The time when the promise will be realized.
%  (slot valid-at (type INTEGER))
%  ; Indicate if the promise should be retracted with its source goal
%  (slot do-not-invalidate (type SYMBOL)
%                          (default FALSE)))
%\end{lstlisting}

\subsection{Using Promises for Planning}
With promises, an agent is able to formulate a goal in expectation of another agent's
actions. However, promises also need to be considered when expanding a goal with a PDDL
planner. To do so, a promise is translated into a
\acfiu{TIL}~\cite{edelkampPDDL2LanguageClassical2004}.  Similar to a promise, a \ac{TIL} states that
some literal will become true at some point in the future, e.g.,
\lstinline|(at 5 (robot-at M1)| states that
\lstinline|(robot-at M1)| will become true in 5 time steps.
%Thus,
%to translate a promise into a \ac{TIL}, we only need to convert the (absolute) promise time into a
%time relative to the planner invocation.
We extended the PDDL plan expansion of the \ac{CX} to
translate promises into \acp{TIL} and to use
\popf~\cite{colesForwardchainingPartialorderPlanning2010}, a temporal PDDL planner that
supports \acp{TIL}.

\begin{figure*}[t]
  \centering
  \begin{subfigure}{\textwidth}
    \scalebox{0.5}{\input{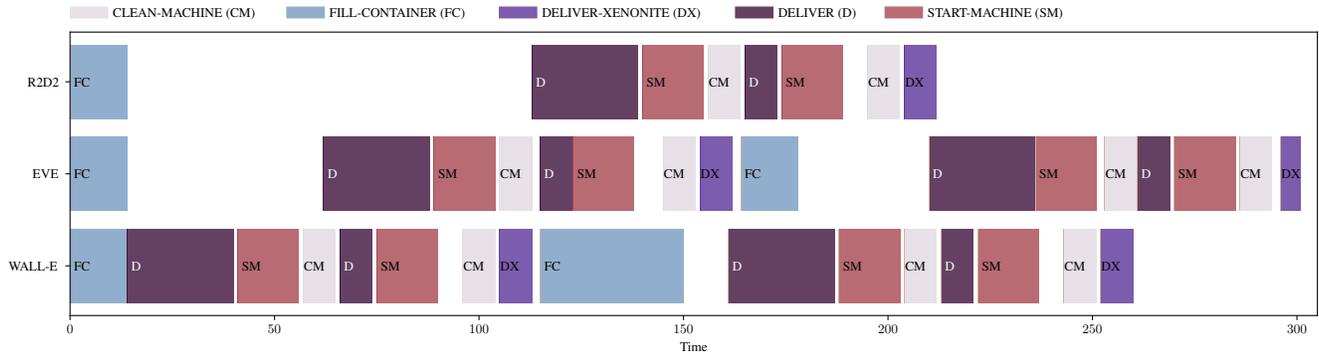}}
    \caption{A run without promises.
      Each goal formulation must wait until the previous goal has finished.
    }
    \label{fig:xenonite-baseline}
  \end{subfigure}
  \begin{subfigure}{\textwidth}
    \scalebox{0.5}{\input{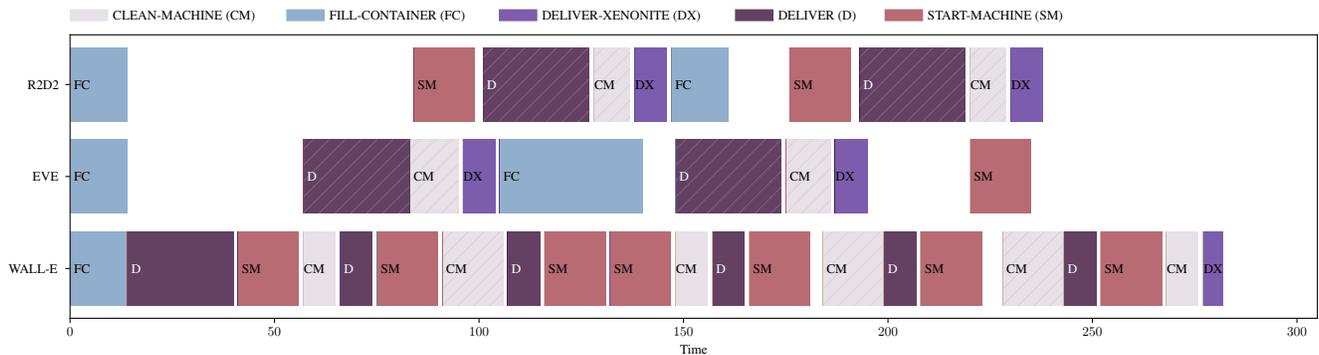}}
    \caption{
      A run with promises.
      The striped goals have been formulated based on promises.
      Compared to the run without promises, the promise-dependent goals (e.g., the first
      \textsc{CleanMachine} goal by \eve) are dispatched earlier, because a future effect of another goal 
      (i.e. the second \textsc{StartMachine} goal by \walle)
      is considered for goal formulation and expansion, thus leading to an overall better
      performance.
      }
    \label{fig:xenonite-promises}
  \end{subfigure}
  \caption{A comparison of two exemplary runs in the Xenonite domain.}
  \label{fig:xenonite}
\end{figure*}

\subsection{Goal Execution with Promises}

When executing a goal that depends on promises (i.e., it has been formulated based on promises), it is necessary to check whether the
promise has actually been realized. To do so, we build on top of the execution engine of the
\ac{CX} \cite{niemuellerCLIPSbasedExecutionPDDL2018}. For each plan action, the \ac{CX} continuously
checks the action's precondition and only starts executing the action once the precondition
is satisfied. While the goal's preconditions can be satisfied (in parts or fully) by
relying on promises, they are not applied to action preconditions.
Thus, the action execution blocks until the precondition is satisfied. This
ensures that actual interactions in the world only operate under the assumption of correct
and current world information. Otherwise, actions might fail or lead to unexpected behavior.   %, depending on the context.

Let us revise our examples from \textsc{Scenario 1} and \textsc{Secenario 2} by introducing
promises, but ignoring the role of goal resources for now. Similarly to the previous scenarios, 
\walle formulates, selects and
executes a goal of class \textsc{StartMachine} on machine \texttt{M1}. With promises, 
the expected outcome of that goal (\lstinline|machine-in-state(M1, READY)|) will 
be announced to the other agents when the goal gets dispatched ($t_1$). \rd now uses
the provided information to formulate the goal \textsc{CleanMachine} ahead of time, even though
the effects of \textsc{StartMachine} have not been applied to the world yet. This behavior is shown in 
\textsc{Scenario 3} of Figure~\ref{fig:timeline-promises}. The first action of the goal's plan (\lstinline|move|) 
can be executed directly. However, the precondition of 
action \lstinline|collect-processite| is not yet satisifed, since the action 
\lstinline|start-machine| is still active on \walle until $t_2$. The executive of \rd notices this
and \lstinline|collect-processite| is pending until the effects have been applied. Once
the effects are applied, \rd can start executing its action.
An example of this behavior occuring in simulation is visualized in Figure~\ref{fig:xenonite}.

\subsection{Execution Monitoring with Promises}
Goals that issue promises might take longer than expected, get stuck, or fail completely, possibly
leading to promises that are never realized. Therefore,
goals that have been formulated based on promises may have actions whose preconditions will never
be satisfied. To deal with this situation, we again rely on execution monitoring.
If a pending action is not executable for a certain amount of time, a timeout occurs and the action
and its corresponding goal are aborted. However, the timeout spans are increased
such that the potentially longer wait times from promise-based execution can be accounted for.

Promises may not cause deadlocks, as a deadlock may only appear if two
goals depend on each other w.r.t. promises. However, this is not possible,
as promises are used to formulate a goal, but only are issued when
the goal is dispatched. Thus, cycles of promises are not possible.

Recall our previous example \textsc{Scenario 3}. Action \lstinline|collect-processite| 
remains in the state pending until \walle finishes the execution of action
\lstinline|start-machine| and its effects are applied. Should \walle not complete the action
(correctly), execution monitoring will detect the timeout on the side of \rd and will trigger
new goal formulation. If the promised-from time has elapsed, the promises
 will not be considered by \rd anymore when formulating new goals, leading to a different objective
 to be chosen by the agent. Should \walle manage to fulfill its goal, then the effects will be applied, and
\rd's original goal's preconditions are satisfied through the actual world state.

\subsection{Resource Locks with Promises}
In the \ac{CX}, resource locks at the goal level are used to coordinate conflicting goals.
Naturally, goals that are formulated on promises often rely on some of the same resources that are
required by the goal that issued the promise. A goal that fills a machine requires the machine as a
resource. Another goal that is promise-dependent and operates on the same machine might require it
as a resoure as well. In this scenario, the second goal may be formulated but would immediately be
rejected, as the required resource is still held by the promising goal.  To resolve this issue,
promise-dependent goals will delay the resource acquisition for any resource that is currently
held by the promising goal. As soon as the promising goal is finished and therefore
the resource is released, the resource is acquired by the promise-dependent goal. To make sure that
there is no conflict between two different promise-dependent goals, for each such delayed resource,
a \emph{promise resource} is acquired, which is simply the main resource prefixed with
\texttt{promised-}. Effectively, a promise-dependent goal first
acquires the promise resource, then, as soon as the promising goal releases the resource, the
promise-dependent goal acquires the main resource, and then releases the promise resource.

To illustrate the mechanism, let us once again consider the example from \textsc{Scenario 3}.
When dispatched, the goal \textsc{StartMachine}
holds the resource \texttt{M1} and \rd formulates the goal \textsc{CleanMachine} based on \walle's
promise. Since the goal operates on the same machine, it needs to acquire the same goal resource. As
the goal is promise-dependent, it first acquires the resource \texttt{promised-M1}. This resource is
currently held by no other agent, thus it can be acquired and the goal dispatched. \rd first
executes the action \texttt{move(BASE, M1)}. As before, the next action
\texttt{collect-processite(M1)} remains pending until \walle has completed its goal. At this point,
\walle releases the resource \texttt{M1}, which is then acquired by \rd.  Should this resource
handover fail, e.g., because the goal \textsc{StartMachine} by \walle never releases its resources,
the action eventually times out and \rd's goal is aborted, as shown in \textsc{Scenario 4} of
Figure~\ref{fig:timeline-promises}.

\section{Evaluation}\label{sec:evaluation}
We evaluate our prototypical implementation\footnote{\url{https://doi.org/10.5281/zenodo.6610426}} of promises for multi-agent cooperation as an extension
of the \ac{CX} in the same simplified production logistics scenario
that we used as a running example throughout this paper. In our evaluation scenario, three robots were
tasked with filling $n=5$ containers with the materials Xenonite and to deliver it to a
storage area. The number of containers was intentionally chosen to be larger than the
number of robots to simulate an oversubscription problem and highlight the effects of
better cooperation on task efficiency. The three robots first collect raw materials and then 
refine them stepwise by passing them through a series of two machines in sequence.
%After each operation, the robots return to the base.
Once all containers are filled and stored, the objective is completed.

%\begin{figure}[ht]
%  \begin{subfigure}{\columnwidth}
%  \scalebox{0.25}{
%    \sfinput{figures/xenonite-no-promises.pgf}
%    }
%\caption{The three robots without promises. A each robot is not aware of the other robots'
%intentions, it can only formulate its goals once the other robot has finished its goal, leading to
%large gaps.}
%\label{fig:xenonite-no-promises}
%\end{subfigure}
%\begin{subfigure}{\columnwidth}
%  \scalebox{0.25}{
%    \sfinput{figures/xenonite-promises.pgf}
%    }
%\caption{The interaction between the three robots with promises. The robots can use promises to
%formulate and dispatch goals early.}
%\label{fig:xenonite-promises}
%\end{subfigure}
%\caption{A comparison of the interaction between the three robots from the Xenonite domain with and
%  without promises.}
%  \label{fig:xenonite-comparison}
%\end{figure}

We compare the performance of the three robots using \popf to expand goals.
Figure \ref{fig:timeline-promises} shows how promises can lead to faster execution of goals by starting
a goal (which always starts with a move action) while the goal's precondition (e.g., that the machine is in
a certain state) is not fulfilled yet.
Figure~\ref{fig:xenonite} shows exemplary runs of three
robots producing and delivering 5 containers of Xenonite. The expected behavior as highlighted
in Figure~\ref{fig:timeline-promises} can indeed be observed in the simulation runs.

To compare the performance, we simulated each scenario five times.
%Because of the small 
%deviation in operational time per run, this was deemed sufficient.
All experiments were carried out on an Intel Core i7 1165G7 machine with \SI{32}{\giga\byte} of RAM.
Without promises, the three robots needed \SI{303}{\sec} to fill and store all containers,
compared to \SI{284.4 \pm 0.55}{\sec} with promises.
%Our evaluation, using with $n=5$, resulted in static execution times for the
%scenario. With promises enabled, the task was completed in 176s. Without promises 190s was needed.
At each call, \popf needed less than \SI{1}{\sec} to generate a plan for the given goal.
This shows, at least in this simplified scenario, that promises lead to more effective
collaboration, and that the planner can make use of promises when expanding a goal.

\section{Conclusion}\label{sec:conclusion}
In multi-agent goal reasoning scenarios, effective collaboration is often difficult, as one agent is
unaware of the goals pursued by the other agents. We have introduced \emph{promises}, a method for
intention sharing in a goal reasoning framework. When dispatching a goal, each agent promises a set of 
literals that will be true at some future timepoint, which can be used by another agent to formulate and
dispatch goals early. We have described how promises can be defined based on \emph{goal operators},
which describe goals similar to how action operators describe action instances. By translating
promises to timed initial literals, we allow the PDDL planner to make use of promises
during goal expansion. The evaluation showed that using promises with our prototypical
implementation improved the performance of a team of three robots in a simplified logistics
scenario. For future work, we plan to use and evaluate promises in a real-world scenario from the
\acf{RCLL}~\cite{niemuellerPlanningCompetitionLogistics2016} and compare it against our distributed
multi-agent goal reasoning system~\cite{hofmannMultiagentGoalReasoning2021}.

\FloatBarrier

\appendix

\section*{Acknowledgements}

This work is funded by the Deutsche Forschungsgemeinschaft (DFG, German Research Foundation) --
2236/1, the EU ICT-48 2020 project TAILOR (No.~952215),
and Germany's Excellence Strategy -- EXC-2023 Internet of Production --
390621612.
{ \small
\bibliography{kbsg-promises}
}

%\section{Acknowledgements}

\end{document}

%% file: clips.tex
\definecolor{Mulberry}{cmyk}{0.34,0.90,0,0.02}
\definecolor{BrickRed}{cmyk}{0,0.89,0.94,0.28}

\lstdefinelanguage{CLIPS}{
  alsoletter={?=-<>*\$},
  keywordstyle=\color{Mulberry!80!black}\bfseries,
  keywords=[2]{deffunction, deftemplate, defrule, deffacts, defgeneric,
    defmodule, defadvice, defglobal, defmethod, definstance, defclass},
  keywordstyle=[2]\color{BrickRed!70!blue}\bfseries,
  keywords=[3]{slot, multislot, type, default, default-dynamic,
                      extends, crlf, range, nil, if, then, else, while,
                      not, or, switch, case, and, reset,
                      assert, test, declare, salience, return, bind, modify,
                      retract, explicit, unique, node-index-hash,
                      halt, printout, =>, <-},
  keywordstyle=[3]\color{darkgray}\bfseries,
  keywords=[4]{subsetp,progn, progn$, not, node-index-hash, create$,
    append$, length$, printout, gensym, gensym*},%
  keywordstyle=[4]\color{gray}\bfseries,
  keywords=[5]{domain-predicate, domain-fact, domain-precondition, domain-atomic-precondition, domain-operator, domain-operator-parameter, domain-effect, plan, plan-action},
  keywordstyle=[5]\color{darkgray}\bfseries,
  sensitive=false,
  comment=[l]{;},
  commentstyle=\color{purple}\ttfamily,
  stringstyle=\color{red}\ttfamily,
  morestring=[b]"
}

\lstdefinestyle{CLIPS}
{
  language=CLIPS,
  basicstyle=\footnotesize\ttfamily\vspace{0.2cm},
  breaklines=true,
  showstringspaces=false,
  frame=lines,
  belowcaptionskip=-8pt,
  emphstyle=\itshape,
  numbers=left,
  stepnumber=1,
  backgroundcolor=\color{gray!3},
  rulecolor=\color{gray!80},
  fillcolor=\color{gray!3},
  framexleftmargin=16pt,
  xleftmargin=16pt,
  stringstyle=\color{BrickRed},
  commentstyle=\color{BrickRed},
  escapechar=\%
}

\lstdefinestyle{SmallCLIPS}{
  style=CLIPS,
  basicstyle=\ttfamily\footnotesize,
  numbersep=6pt,
  captionpos=b
}
\lstdefinestyle{ReallySmallCLIPS}{
  style=CLIPS,
  basicstyle=\ttfamily\scriptsize,
  numbersep=5pt,
  captionpos=b
}

\lstdefinestyle{ReallySmallCLIPSNoFrame}{
  style=CLIPS,
  basicstyle=\ttfamily\scriptsize,
  numbersep=5pt,
  frame=none,
  backgroundcolor=\color{white},
  framextopmargin=0pt,
  framexbottommargin=0pt
}

\lstdefinestyle{SuperSmallCLIPSNoFrame}{
  style=CLIPS,
  basicstyle=\ttfamily\fontsize{10pt}{10pt}\selectfont,
  numbers=none,
  frame=none,
  backgroundcolor=\color{white},
  framextopmargin=0pt,
  framexbottommargin=0pt,
  framexleftmargin=-2pt, xleftmargin=-2pt,
}

\lstdefinestyle{SmallCLIPSNoFrame}{
  style=CLIPS,
  basicstyle=\ttfamily\footnotesize,
  numbersep=5pt,
  frame=none,
  backgroundcolor=\color{white},
  framextopmargin=0pt,
  framexbottommargin=0pt
}

\lstdefinestyle{SmallCLIPSFloat}{
  style=SmallCLIPS,
  float,
}

\lstdefinestyle{ReallySmallCLIPSFloat}{
  style=ReallySmallCLIPS,
  float,
}

%% file: abstract.txt
Reasoning and planning for mobile robots is a challenging problem, as the world
evolves over time and thus the robot's goals may change. One technique to
tackle this problem is goal reasoning, where the agent not only reasons about
its actions, but also about which goals to pursue. While goal reasoning for
single agents has been researched extensively, distributed, multi-agent goal
reasoning comes with additional challenges, especially in a distributed setting.
In such a context, some form of coordination is necessary to allow for cooperative
behavior. Previous goal reasoning approaches share the agent's world model
with the other agents, which already enables basic cooperation.
However, the agent's goals, and thus its intentions, are typically not shared.

In this paper, we present a method to tackle this limitation. Extending an
existing goal reasoning framework, we propose enabling cooperative behavior
between multiple agents through promises, where an agent may promise that certain
facts will be true at some point in the future. Sharing these promises allows
other agents to not only consider the current state of the world, but also the
intentions of other agents when deciding on which goal to pursue next.  We describe
how promises can be incorporated into the goal life cycle, a commonly used goal
refinement mechanism. We then show how promises can be used when planning for a particular
goal by connecting them to timed initial literals (TILs) from PDDL planning.
Finally, we evaluate our prototypical implementation in a simplified logistics
scenario.